%% file: main.tex
\documentclass[acmtog,authorversion]{acmart}

\usepackage{booktabs} % For formal tables
\usepackage[ruled]{algorithm2e} % For algorithms

\SetAlFnt{\small}
\SetAlCapFnt{\small}
\SetAlCapNameFnt{\small}
\SetAlCapHSkip{0pt}
\IncMargin{-\parindent}

\setcopyright{none}

\usepackage{caption}
\usepackage{subcaption}
\usepackage{fix2col}
\usepackage{xcolor}
\usepackage{dsfont}
\usepackage{xfrac}

\begin{document}
% Title portion
\title{Real-Time User-Guided Image Colorization with Learned Deep Priors} 
\author{Richard Zhang*}
\orcid{0000-0002-1522-2381}
\affiliation{%
  \institution{University of California, Berkeley}
%   \institution{}
  \department{Electrical Engineering and Computer Science}
  \city{Berkeley}
  \state{CA}
  \postcode{94720}
}
\author{Jun-Yan Zhu*}
\affiliation{%
  \institution{University of California, Berkeley}
%   \institution{}
  \department{Electrical Engineering and Computer Science}
  \city{Berkeley}
  \state{CA}
  \postcode{94720}
}
\author{Phillip Isola}
\affiliation{%
  \institution{University of California, Berkeley}
%   \institution{}
  \department{Electrical Engineering and Computer Science}
  \city{Berkeley}
  \state{CA}
  \postcode{94720}
}
\author{Xinyang Geng}
\affiliation{%
  \institution{University of California, Berkeley}
%   \institution{}
  \department{Electrical Engineering and Computer Science}
  \city{Berkeley}
  \state{CA}
  \postcode{94720}
}
\author{Angela S. Lin}
\affiliation{%
  \institution{University of California, Berkeley}
%   \institution{}
  \department{Electrical Engineering and Computer Science}
  \city{Berkeley}
  \state{CA}
  \postcode{94720}
}
\author{Tianhe Yu}
\affiliation{%
  \institution{University of California, Berkeley}
%   \institution{}
  \department{Electrical Engineering and Computer Science}
  \city{Berkeley}
  \state{CA}
  \postcode{94720}
}
\author{Alexei A. Efros}
\affiliation{%
  \institution{University of California, Berkeley}
%   \institution{}
  \department{Electrical Engineering and Computer Science}
  \city{Berkeley}
  \state{CA}
  \postcode{94720}
}

\begin{teaserfigure}
   \centering
\includegraphics[width=1.\textwidth]{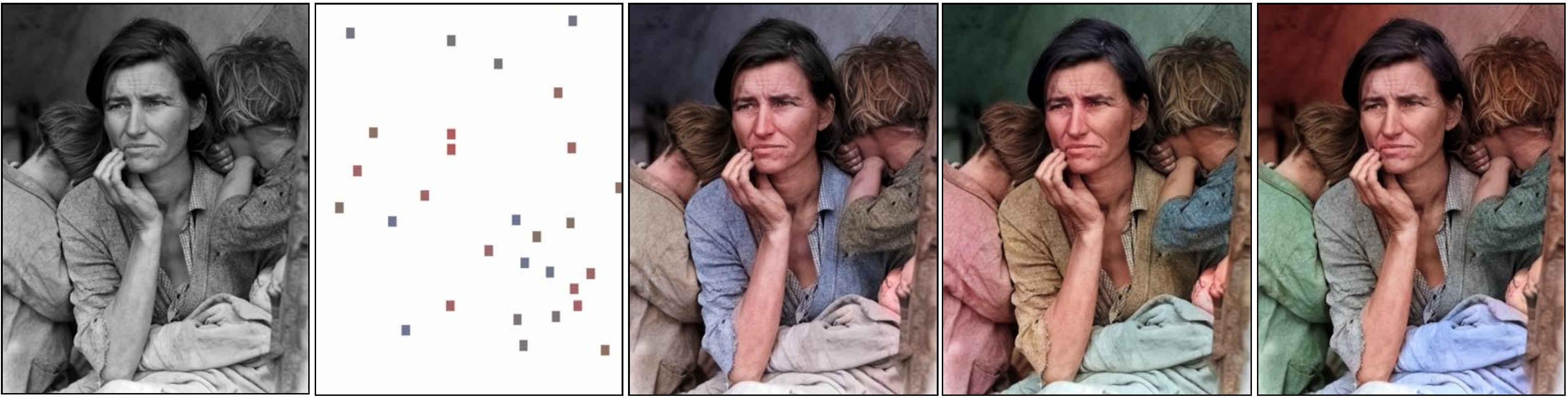}
\vspace{-6mm}
\caption{Our proposed method colorizes a grayscale image (left), guided by sparse user inputs (second), in real-time, providing the capability for quickly generating multiple plausible colorizations (middle to right). Photograph of \textit{Migrant Mother} by Dorothea Lange, 1936 (Public Domain).}
\end{teaserfigure}

\begin{abstract}
We propose a deep learning approach for user-guided image colorization. The system directly maps a grayscale image, along with sparse, local user ``hints" to an output colorization with a Convolutional Neural Network (CNN). Rather than using hand-defined rules, the network propagates user edits by fusing low-level cues along with high-level semantic information, \textit{learned from large-scale data}. We train on a million images, with simulated user inputs. To guide the user towards efficient input selection, the system recommends likely colors based on the input image and current user inputs. The colorization is performed in a single feed-forward pass, enabling real-time use. Even with randomly simulated user inputs, we show that the proposed system helps novice users quickly create realistic colorizations, and offers large improvements in colorization quality with just a minute of use. In addition, we demonstrate that the framework can incorporate other user ``hints" to the desired colorization, showing an application to color histogram transfer. Our code and models are available at \url{https://richzhang.github.io/ideepcolor}.
\end{abstract}


\acmJournal{TOG}
\acmYear{2017}\acmVolume{36}\acmNumber{4}\acmArticle{119}\acmMonth{7}
\acmDOI{http://dx.doi.org/10.1145/3072959.3073703}


\maketitle

\input{sections/1_intro}
\input{sections/2_related}

\input{sections/3_methods}

\input{sections/4_experiments}
\input{sections/5_conclusions}
\input{sections/6_acknowledgements}
\input{sections/7_changelog}

\bibliography{main}

\end{document}

%% file: sections/1_intro.tex
% \blfootnote{\small{* indicates equal contribution}}
\section{Introduction}

There is something uniquely and powerfully satisfying about the simple act of adding color to black and white imagery. Whether as a way of rekindling old, dormant memories or expressing artistic creativity, people continue to be fascinated by colorization.  From remastering classic black and white films, to the enduring popularity of coloring books for all ages, to the surprising enthusiasm for various (often not very good) automatic colorization bots online\footnote{e.g., \url{http://demos.algorithmia.com/colorize-photos/} \\ * indicates equal contribution}, this topic continues to fascinate the public. 

In computer graphics, two broad approaches to image colorization exist: user-guided edit propagation and data-driven automatic colorization. In the first paradigm, popularized by the seminal work of Levin et al.~\shortcite{levin2004colorization}, a user draws colored strokes over a grayscale image. An optimization procedure then generates a colorized image that matches the user's scribbles, while also adhering to hand-defined image priors, such as piecewise smoothness. These methods can achieve impressive results but often require intensive user interaction (sometimes over fifty strokes), as each differently colored image region must be explicitly indicated by the user. Because the system purely relies on user inputs for colors, even regions with little color uncertainty, such as green vegetation, need to be specified. Less obviously, even if a user knows what general color an object should take on, it can be surprisingly difficult to select the exact desired natural chrominance.

To address these limitations, researchers have also explored more data-driven colorization methods. These methods colorize a grayscale photo in one of two ways: either by matching it to an exemplar color image in a database and non-parametrically ``stealing'' colors from that photo, an idea going back to Image Analogies~\citep{hertzmann2001image}, or by learning parametric mappings from grayscale to color from large-scale image data.
% The most recent methods in this paradigm use deep networks and are fully automatic~\cite{iizuka2016let, larsson2016learning, zhang2016colorful}.
The most recent methods in this paradigm proposed by Iizuka et al.~\shortcite{iizuka2016let}, Larsson et al.~\shortcite{larsson2016learning}, and Zhang et al.~\shortcite{zhang2016colorful}, use deep networks and are fully automatic. Although this makes colorizing a new photo cheap and easy, the results often contain incorrect colors and obvious artifacts. More fundamentally, the color of an object, such as a t-shirt, is often inherently ambiguous -- it could be blue, red, or green. Current automatic methods aim to choose a single colorization, and do not allow a user to specify their preference for a plausible, or perhaps artistic, alternative.

Might we be able to get the best of both worlds, leveraging large-scale data to learn priors about natural color imagery, while at the same time incorporating user control from traditional edit propagation frameworks? We propose to train a CNN to directly map grayscale images, along with sparse user inputs, to an output colorization. During training, we randomly simulate user inputs, allowing us to bypass the difficulty of collecting user interactions. Though our network is trained with ground truth natural images, the network can colorize objects with different, or even unlikely colorizations, if desired.

Most traditional tools in interactive graphics are defined either procedurally -- e.g., as a designed image filter -- or as constraints applied in a hand-engineered optimization framework. The behavior of the tool is therefore fully specified by human fiat. This approach is fundamentally limited by the skill of engineers to design complex operations/constraints that actually accomplish what is intended of them. Our approach differs in that the effect of interaction is \emph{learned}. Through learning, the algorithm may come up with a more powerful procedure for translating user edits to colorized results than would be feasible by human design.

Our contribution are as follows: (1) We \textit{end-to-end learn} how to propagate sparse user points from large-scale data, by training a deep network to directly predict the mapping from grayscale image and user points to full color image. (2) To guide the user toward making informed decisions, we provide a data-driven color palette, which suggests the most probable colors at any given location. (3) We run a study, showing that even given minimal training with our interface and limited time to colorize an image (1 min), novice users can quickly learn to produce colorizations that can often fool real human judges in a real vs. fake test. (4) Though our system is trained on natural images, it can also generate unusual colorizations. (5) We demonstrate that this framework is not limited to user points, and can, in principle, be trained with any statistic of the output, for example, global color distribution or average image saturation.

%% file: sections/2_related.tex
\begin{figure*}[t!]
\centering
\includegraphics[width=1.0\hsize]{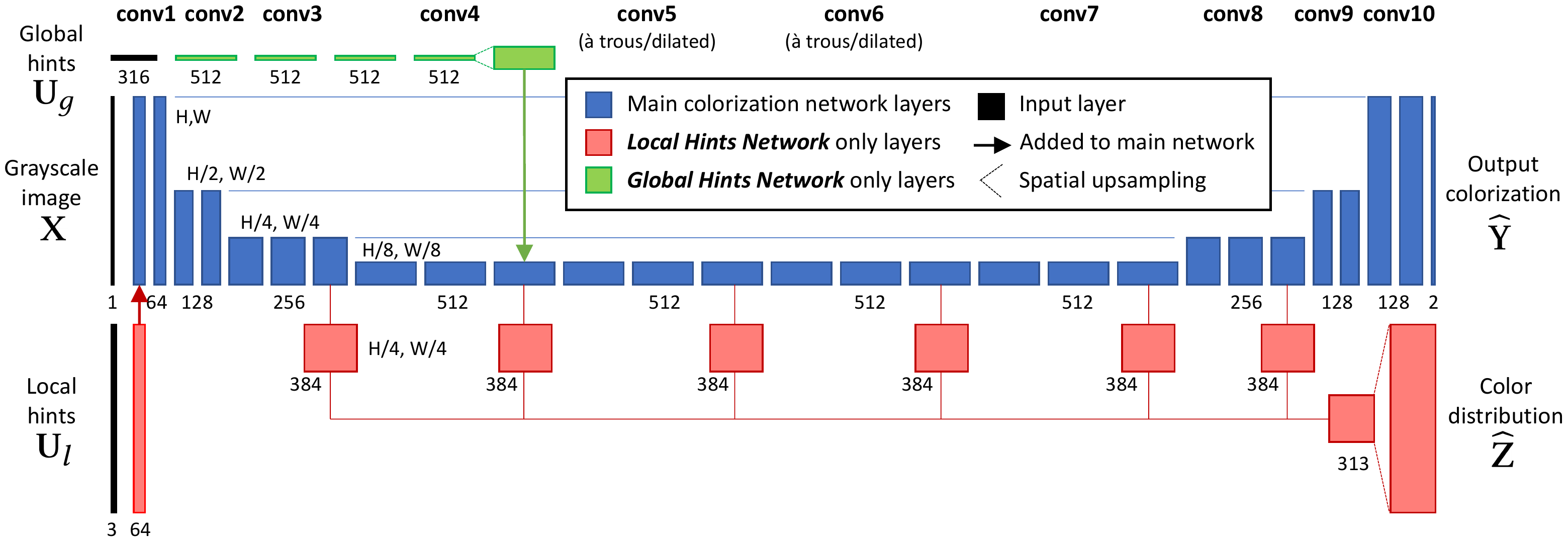}
\vspace{-5mm}
\caption{\textbf{Network architecture} We train two variants of the user interaction colorization network. Both variants use the blue layers for predicting a colorization. The \textbf{Local Hints Network} also uses red layers to (a) incorporate user points $\mathbf{U}_l$ and (b) predict a color distribution $\mathbf{\widehat{Z}}$. The \textbf{Global Hints Network} uses the green layers, which transforms global input $\mathbf{U}_g$ by $1\times 1$ \texttt{conv} layers, and adds the result into the main colorization network. Each box represents a \texttt{conv} layer, with vertical dimension indicating feature map spatial resolution, and horizontal dimension indicating number of channels. Changes in resolution are achieved through subsampling and upsampling operations. In the main network, when resolution is decreased, the number of feature channels are doubled. Shortcut connections are added to upsampling convolution layers.}
\label{fig:network}
\vspace{-2mm}
\end{figure*}

\section{Related Work}

\paragraph{User-guided colorization} Prior interactive colorization work focused on local control, such as user strokes \cite{levin2004colorization,huang2005adaptive}. Because the strokes are propagated using low-level similarity metrics, such as spatial offset and intensity difference, numerous user edits are typically required to achieve realistic results. To reduce user efforts, later methods focused on designing better similarity metrics \cite{qu2006manga,luan2007natural} and utilizing long-range connections \cite{an2008appprop,xu2009efficient}. Learning machinery, such as boosting \cite{li2008scribbleboost}, local linear embeddings \cite{chen2012manifold}, feature discrimination \cite{xu2013sparse}, and more recently, neural networks \cite{endo2016deepprop}, have been proposed to automatically learn similarity between pixels given user strokes and input images. In addition to local control, varying the color theme~\cite{wang2010data,li2015image} and color palette~\cite{chang2015palette} are popular methods of expressive global control. We show that we can integrate global hints to our network and control colorization results by altering the color distribution and average saturation (see Section~\ref{sec:globalhintsnet}). Concurrently, Sangkloy et al.~\shortcite{sangkloy2017scribbler} developed a system to translate sketches to real images, with support for user color strokes, while PaintsChainer~\shortcite{paintschainer} and Frans ~\shortcite{frans2017outline} have developed open-source interactive online applications for line-drawing colorization. 

\paragraph{Automatic colorization} 
Early semi-automatic methods~\cite{welsh2002transferring,irony2005colorization,liu2008intrinsic,chia2011semantic,gupta2012image} utilize an example-based approach that transfers color statistics from a reference image or multiple images~\cite{morimoto2009automatic,liu2014autostyle} to the input grayscale image with techniques such as color transfer~\cite{Reinhard2001color} and image analogies~\cite{hertzmann2001image}. These methods work remarkably well when the input and the reference share similar content. However, finding reference images is time-consuming and can be challenging for rare objects or complex scenes, even when using semi-automatic retrieval methods~\cite{chia2011semantic}. In addition, some algorithms ~\cite{irony2005colorization,chia2011semantic} involve tedious manual efforts on defining corresponding regions between images.

Recently, fully automatic methods \cite{deshpande2015learning,cheng2015deep,iizuka2016let,zhang2016colorful,larsson2016learning,isola2016image} have been proposed. The recent methods from train CNNs~\cite{lecun1998gradient} on large-scale image collections~\cite{russakovsky2015imagenet,zhou2014learning} to directly map grayscale images to output colors. The networks can learn to combine low and high-level cues to perform colorization, and have been shown to produce realistic results, as determined by human judgments \cite{zhang2016colorful}. However, these approaches aim to produce a \textit{single} plausible result, even though colorization is intrinsically an ill-posed problem with multi-modal uncertainty~\cite{charpiat2008automatic}. Larsson et al.~\shortcite{larsson2016learning} provide some post-hoc control through globally biasing the hue, or by matching global statistics to a target histogram. Our work addresses this problem by learning to integrate input hints in an end-to-end manner.

\paragraph{Deep semantic image editing}
Deep neural networks~\cite{krizhevsky2012imagenet} excel at extracting rich semantics from images, from middle-level concepts like material~\cite{bell2015material,wang20164d} and segmentation~\cite{xie2015holistically}, to high-level knowledge such as objects~\cite{girshick2014rich} and scene categories~\cite{zhou2014learning}. All of this information could potentially benefit semantic image editing, i.e. changing the high-level visual content with minimal user interaction. Recently, neural networks have shown impressive results for various image processing tasks, such as photo enhancement~\cite{yan2016automatic}, sketch simplification~\cite{simo2016learning}, style transfer~\cite{gatys2016image,selim2016painting}, inpainting~\cite{pathakCVPR16context}, image blending~\cite{zhu2015learning} and denoising~\cite{gharbi2016deep}. Most of these works built image filtering pipelines and trained networks that produce a filtered version of the input image with different low-level local details. However, none of these methods allowed dramatic, high-level modification of the visual appearance, nor do they provide diverse outputs in a user controllable fashion. On the contrary, we train a network that takes an input image as well as minimal user guidance and produces global changes in the image with a few clicks. 
Barnes et al.~\shortcite{barnes2009patchmatch} emphasize that control and interactivity are key to image editing, because user intervention not only can correct errors, but can also help explore the vast design space of creative image manipulation. We incorporate this concept into an intuitive interface that provides expressive controls as well as real-time feedback. Zhu et al. ~\shortcite{zhu2016generative} provided an interactive deep image synthesis interface that builds on an image prior learned by a deep generative network. Xu et al.~\shortcite{xu2016deep} train a deep network for interactive object segmentation. Isola et al.~\shortcite{isola2016image} and Sangkloy et al.~\shortcite{sangkloy2017scribbler} train networks to generate images from sketches, using synthetic sketches generated by edge detection algorithms for training data.

%% file: sections/3_methods.tex
\section{Methods}

We train a deep network to predict the color of an image, given the grayscale version and user inputs. In Section \ref{sec:learncolor}, we describe the objective of the network. We then describe the two variants of our system (i) the \textbf{Local Hints Network} in Section \ref{sec:localhintsnet}, which uses sparse user points, and (ii) the \textbf{Global Hints Network} in Section \ref{sec:globalhintsnet}, which uses global statistics. In Section \ref{sec:net-arch}, we define our network architecture.

\subsection{Learning to Colorize}
\label{sec:learncolor}

The inputs to our system are a grayscale image $\mathbf{X}\in \mathds{R}^{H\times W \times 1}$, along with an input user tensor $\mathbf{U}$. The grayscale image is the $L$, or lightness in the CIE $Lab$ color space, channel. The output of the system is $\mathbf{\widehat{Y}} \in \mathds{R}^{H\times W \times 2}$, the estimate of the $ab$ color channels of the image. The mapping is learned with a CNN $\mathcal{F}$, parameterized by $\theta$, with the network architecture specified in Section \ref{sec:net-arch} and shown in Figure \ref{fig:network}. We train the network to minimize the objective function in Equation \ref{eqn:minF}, across $\mathcal{D}$, which represents a dataset of grayscale images, user inputs, and desired output colorizations. Loss function $\mathcal{L}$ describes how close the network output is to the ground truth.

\vspace{-3mm}
\begin{equation}
\mathcal{\theta}^{*} = \arg\min_{\mathcal{\theta}} \mathds{E}_{\mathbf{X},\mathbf{U},\mathbf{Y}\sim \mathcal{D}} [ \mathcal{L}(\mathcal{F}(\mathbf{X},\mathbf{U}; \theta),\mathbf{Y}) ] \\
\label{eqn:minF}
\end{equation}
\vspace{-3mm}

We train two variants of our network, with local user hints $\mathbf{U}_l$ and global user hints $\mathbf{U}_g$. During training, the hints are generated by giving the network a ``peek", or projection, of the ground truth color $\mathbf{Y}$ using functions $\mathcal{P}_l$ and $\mathcal{P}_g$, respectively.

% add expectation
\vspace{-3mm}
\begin{equation}
% \begin{split}
\mathbf{U}_l = \mathcal{P}_l(\mathbf{Y}) \hspace{1cm}
\mathbf{U}_g = \mathcal{P}_g(\mathbf{Y})
% \end{split}
\label{eqn:samp}
\end{equation}
\vspace{-3mm}

The minimization problems for the Local and Global Hints Networks are then described below in Equation \ref{eqn:minFvars}. Because we are using functions $\mathcal{P}_l, \mathcal{P}_g$ to synethtically generate user inputs, our dataset only needs to contain grayscale and color images. We use the 1.3M ImageNet dataset~\cite{russakovsky2015imagenet}.

% add expectation
\begin{equation}
\begin{split}
& \mathcal{\theta}^{*}_{l} = \arg\min_{\mathcal{\theta}_{l}} \mathds{E}_{\mathbf{X},\mathbf{Y}\sim \mathcal{D}} [ \mathcal{L}(\mathcal{F}_{l}(\mathbf{X},\mathbf{U}_{l}; \theta_{l}),\mathbf{Y}) ] \\
& \mathcal{\theta}^{*}_{g} = \arg\min_{\mathcal{\theta}_{g}} \mathds{E}_{\mathbf{X},\mathbf{Y}\sim \mathcal{D}} [ \mathcal{L}(\mathcal{F}_{g}(\mathbf{X},\mathbf{U}_{g}; \theta_{g}),\mathbf{Y}) ]
\end{split}
\label{eqn:minFvars}
\end{equation}
\vspace{-4mm}

\begin{figure*}[t!]
\centering
\includegraphics[width=1.\hsize]{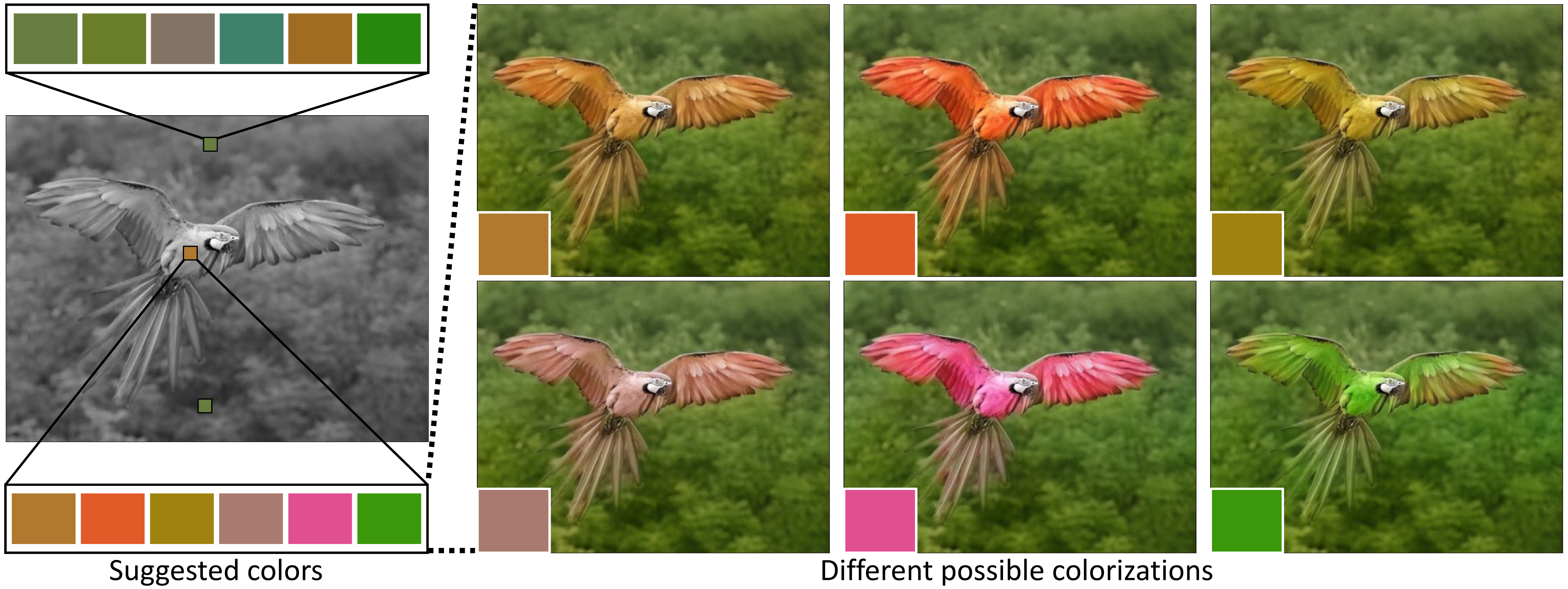}
\caption{\textbf{Suggested Palette} Our interface provides suggested colors for any pixel, sorted by likelihood, based on the predicted color distribution given by our network. In this example, we show first suggested colors on the background vegetation (top palette), sorted by decreasing likelihood. The suggested colors are common colors for vegetation. We also show the top six suggested colors (bottom palette) of a pixel on the image of the bird. On the right, we show the resulting colorizations, based on the user selecting these top six suggested colors. Photograph of blue-and-yellow macaw by Luc Viatour, 2009.}
\vspace{-2mm}
\label{fig:palette}
\end{figure*}

\paragraph{Loss Function} The choice of an appropriate loss function $\mathcal{L}$, which measures network performance and guides learning, requires some consideration. Iizuka et al.~\shortcite{iizuka2016let} use an $\ell_2$ loss. Previous work~\cite{zhang2016colorful,larsson2016learning,charpiat2008automatic} note that this loss is not robust to the inherent multi-modal nature of the problem, and instead use a classification loss, followed by a fixed inference step. Another challenge is the large imbalance in natural image statistics, with many more pixels in desaturated regions of the color gamut. This can often lead to desaturated and dull colorizations. Zhang et al.~\shortcite{zhang2016colorful} use a class-rebalancing step to oversample more colorful portions of the gamut during training. This results in more colorizations which are vibrant and able to fool humans, but at the expense of images which are over-aggressively colorized. In the pix2pix framework, Isola et al.~\shortcite{isola2016image} use an $\ell_1$ regression loss with a Generative Adversarial Network (GAN)~\cite{goodfellow2014generative} term, which can help generate exciting, higher frequency patterns.

However, in our work, we forgo the use of class rebalancing from~\cite{zhang2016colorful} and GAN term from ~\cite{isola2016image} and use a smooth-$\ell_1$ (or Huber) loss, described in Equation~\ref{eqn:huber}. In the Local Hints Network, from a user experience standpoint, we found it more pleasing to start with a conservative colorization and allow the user to inject desired colors, rather than starting with a more vibrant but artifact-prone setting and having the user fix mistakes. Much of the multi-modal ambiguity of the problem is quickly resolved by a few user clicks. In cases where there is ambiguity, the smooth-$\ell_1$ is also a robust estimator \cite{huber1964robust}, which can help avoid the averaging problem. In addition, using a regression loss, described in Equation \ref{eqn:huber} with $\delta=1$, enables us to perform end-to-end learning without a fixed inference step.

\vspace{-2mm}
\begin{equation}
\ell_{\delta}(x,y) = 
%  \begin{cases} 
%   \tfrac{1}{2}x^2 & |x|< \delta \\
%   \delta(|x|-\tfrac{1}{2}\delta) & |x|\geq \delta
%     \end{cases}
\tfrac{1}{2}(x-y)^2 \mathds{1}_{\big\{ |x-y|< \delta \big\} }
+ \delta(|x-y|-\tfrac{1}{2}\delta) \mathds{1}_{\big\{ |x-y|\geq \delta \big\} }
\label{eqn:huber}
\end{equation}
\vspace{-1mm}

The loss function $\ell_\delta$ is evaluated at each pixel and summed together to evaluate the loss $\mathcal{L}$ for a whole image.

\vspace{-2mm}
\begin{equation}
\mathcal{L}(\mathcal{F}(\mathbf{X},\mathbf{U};\theta),\mathbf{Y}) = \sum_{h,w} \sum_{q} \ell_{\delta} \big( \mathcal{F}(\mathbf{X,U};\theta)_{h,w,q},\mathbf{Y}_{h,w,q} \big)
\label{eqn:loss}
\end{equation}

\noindent Next, we describe the specifics of the local and global variants.

\subsection{Local Hints Network}
\label{sec:localhintsnet}

The Local Hints Network uses sparse user points as input. We describe the input, how we simulate user points, and features of our user interface.

% \subsubsection{System Input}
\paragraph{System Input}
\label{sec:lhn-input}
The user points are parameterized as $\mathbf{X}_{ab}\in \mathds{R}^{H\times W \times 2}$, a sparse tensor with $ab$ values for the points provided by the user and $\mathbf{B}_{ab}\in \mathds{B}^{H\times W \times 1}$, a binary mask indicating which points are provided by the user. The mask differentiates unspecified points from user-specified gray points with $(a,b)=0$. Together, the tensors form input tensor $\mathbf{U}_{l}=\{\mathbf{X}_{ab}, \mathbf{B}_{ab}\}\in \mathds{R}^{H\times W \times 3}$.

% \subsubsection{Simulating User Interactions}
\paragraph{Simulating User Interactions} One challenge in training deep networks is collecting training data. While data for automatic colorization is readily available -- any color image can be broken up into its color and grayscale components -- an appropriate mechanism for acquiring user interaction data is far less obvious. Gathering this on a large scale is not only expensive, but also comes with a chicken and egg problem, as user interaction behavior will be dependent on the system performance itself. We bypass this issue by training with synthetically generated user interactions. A concern with this approach is the potential domain gap between the generated data and test-time usage. However, we found that even through \textit{randomly sampling}, we are able to cover the input space adequately and train an effective system.

We sample small patches and reveal the average patch color to the network. For each image, the number of points are drawn from a geometric distribution with $p=\tfrac{1}{8}$. Each point location is sampled from a 2-D Gaussian with $\mu=\tfrac{1}{2}[H,W]^{T}, \Sigma=diag \big( \big[ \big( \tfrac{H}{4} \big) ^{2},\big( \tfrac{W}{4} \big) ^{2} \big] \big)$, as we expect users to more often click on points in the center of the image. The revealed patch size is drawn uniformly from size $1\times 1$ to $9\times 9$, with the average $ab$ within the patch revealed to the network. Lastly, we desire the correct limiting characteristic -- given all of the points by the user, the network should simply copy the colors from the input to the output. To encourage this, we provide the full ground truth color to the image for 1\% of the training instances. Though the network should implicitly learn to copy any provided user points to the output, there is no explicit constraint for the network to do so exactly. Note that these design decisions for projection function $\mathcal{P}_l(\mathbf{Y})$ were initially selected based on intuition, found to work well, but not finely tuned.

% \subsubsection{User interface}
\paragraph{User interface} Our interface consists of a drawing pad, showing user points overlaid on the grayscale input image, a display updating the colorization result in real-time, a data-driven color palette that suggests likely color for a given location (as shown in Figure~\ref{fig:palette}), and a regular $ab$ gamut based on the lightness of the current point. A user is always free to add, move, delete, or change the color of any existing points. Please see our supplemental video for a detailed introduction of our interface, along with several demonstrations. 

% \subsubsection{Data-driven color palette}
\paragraph{Data-driven color palette} Picking a plausible color is an important step towards realistic colorization. Without the proper tools, selecting a color can be difficult for a novice user to intuit. For every pixel, we predict a probability distribution over output colors $\mathbf{\widehat{Z}} \in \mathds{R}^{H\times W \times Q}$, where $Q$ is the number of quantized color bins. We use the parametrization of the CIE $Lab$ color space from Zhang et al.~\shortcite{zhang2016colorful} -- the $ab$ space is divided into $10\times 10$ bins, and the $Q=313$ bins that are in-gamut are kept. The mapping from the input grayscale image and user points to predicted color distribution $\mathbf{\widehat{Z}}$ is learned with network $\mathcal{G}_l$, parametrized by $\psi_l$. Ground truth distribution $\mathbf{Z}$ is encoded from ground truth colors $\mathbf{Y}$ with the soft-encoding scheme from \cite{zhang2016colorful} -- a real $ab$ color value is expressed as a convex combination of its 10 nearest bin centers, weighted by a Gaussian kernel with $\sigma=5$. We use a cross-entropy loss function for every pixel to measure the distance between predicted and ground truth distributions, and sum over all pixels.

\vspace{-2mm}
\begin{equation}
\mathcal{L}_{cl}(\mathcal{G}_l(\mathbf{X,U}_l;\psi_l),\mathbf{Z}) = - \sum_{h,w}\sum_{q} \mathbf{Z}_{h,w,q} \log (\mathcal{G}_l(\mathbf{X,U}_l;\psi_l)_{h,w,q})
\label{eqn:classloss}
\end{equation}

Network $\mathcal{G}_l$ is trained to minimize expected classification loss over the training set. We further describe the network architecture in Section \ref{sec:net-arch}.

\vspace{-2mm}
\begin{equation}
\psi_l^{*} = \arg\min_{\psi_l} \mathds{E}_{\mathbf{X},\mathbf{Y}\sim \mathcal{D}} [ \mathcal{L}_{cl}(\mathcal{G}_l(\mathbf{X,U}_l;\psi_l),\mathbf{Y}) ] \\
\label{eqn:minG}
\end{equation}

To provide discrete color suggestions, we soften the softmax distribution at the queried pixel, to make it less peaky, and perform weighted k-means clustering (with $K=9$) to find modes of the distribution. For example, the system often recommends plausible colors based on the type of object, material and texture, for example, suggesting different shades of green the vegetation in Figure~\ref{fig:palette}. For objects with diverse colors such as a parrot, our system will provide a wide range of suggestions. Once a user selects a suggested color, our system will produce the colorization result in real-time. In Figure~\ref{fig:palette}, we show six possible colorizations based on the different choices for the parrot's feather. The color suggestions are continuously updated as the user adds additional points.

\begin{figure*}[h!]
\vspace{-2mm}
\centering
\includegraphics[width=1.\hsize]{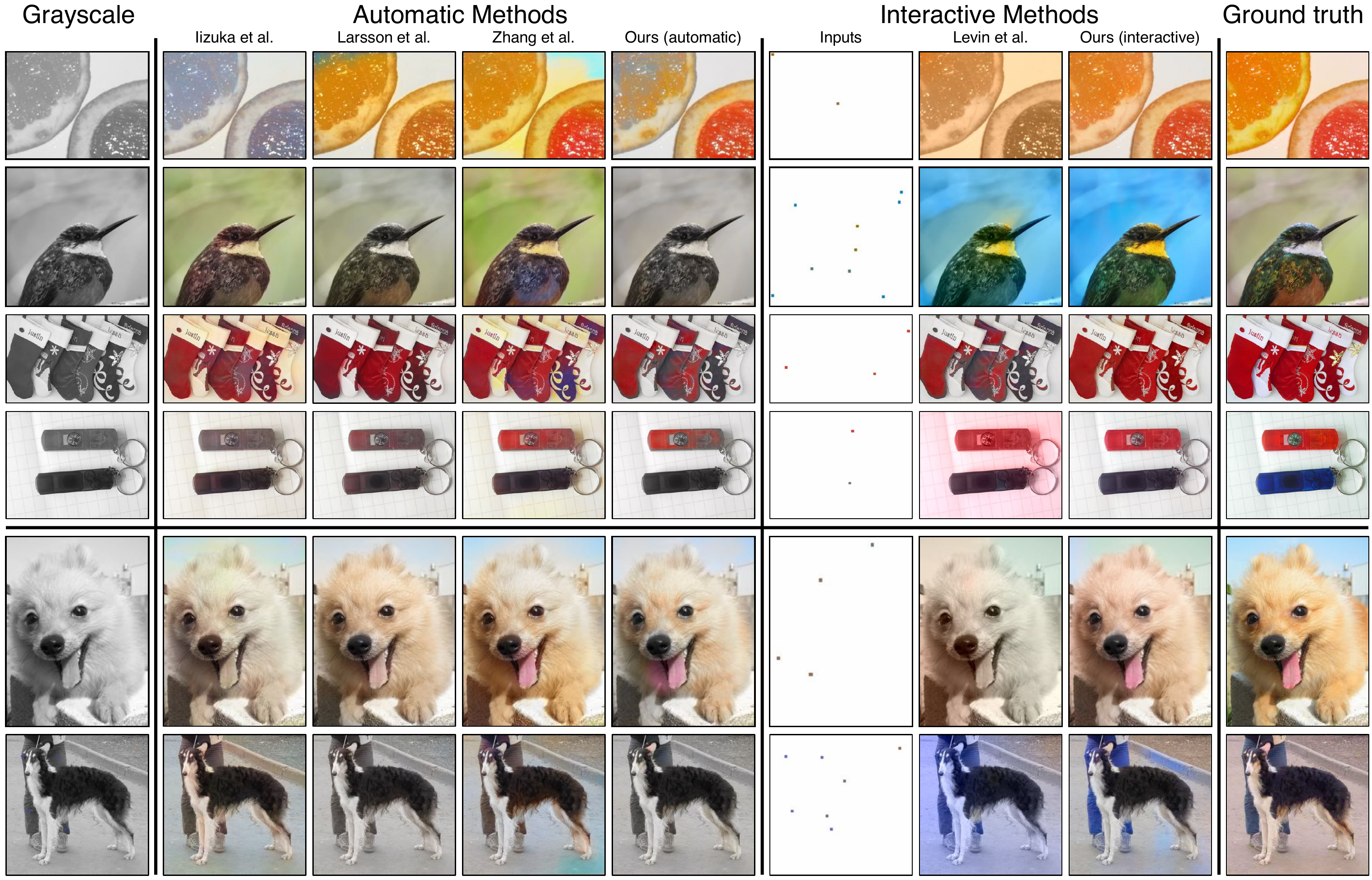}
\vspace{-4mm}
\caption{\textbf{User study results} These results are collected from novice users using our Local Hints Network system for 1 minute for each image, with minimal training. Users were not given the ground truth image, and were instructed to create a ``realistic colorization". The first column shows the grayscale input image. Columns 2-5 show automatic results from previous methods, as well as our system without user points. Column 6 shows input points from a user, collected in 1 minute of time from a novice user. Columns 7-8 show the results from the seminal method of \cite{levin2004colorization} and our model, incorporating user points, on the right. The final column shows the ground truth image (which was not provided to the user). In the selected examples in rows 1-4, our system produces higher quality colorizations given sparse inputs than \cite{levin2004colorization}, and produce nearly photorealistic results given little user interaction. Rows 5-6 show some failures of our system. In row 5, the green color by the user on the top-right is not successfully propagated to the top-left of the image. In row 6, the colors selected on the jeans are propagated to the background, demonstrating undesired non-local effects. All of the user study results are publicly available on \url{https://richzhang.github.io/ideepcolor/}. Images are from the ImageNet dataset~\cite{russakovsky2015imagenet}.}
\label{fig:userstudy_res}
\vspace{-2mm}
\end{figure*}

\begin{figure*}[t!]
\includegraphics[width=1.\hsize]{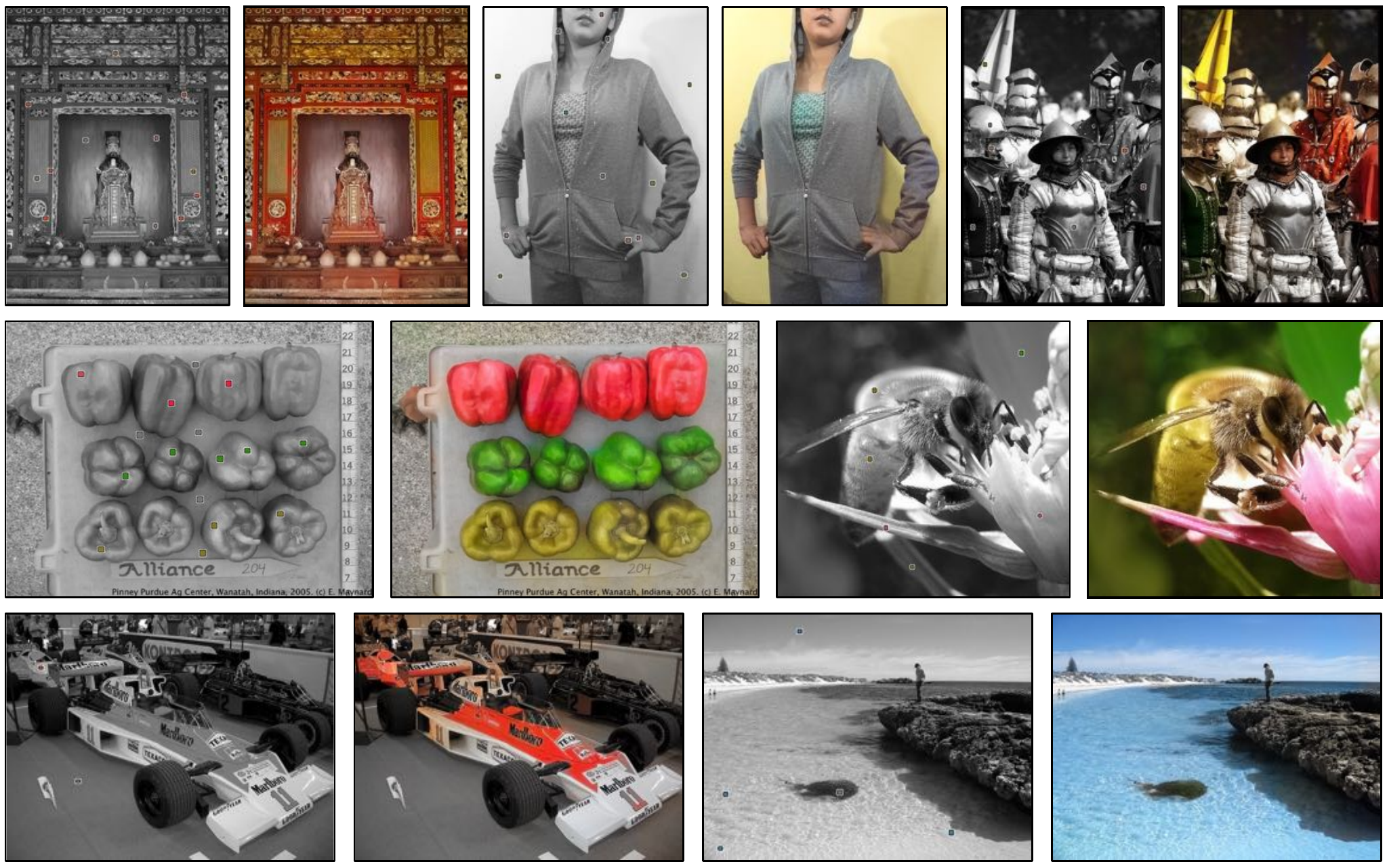}
\vspace{-6mm}
\caption{\textbf{Selected User Study Results} We show grayscale images with user inputs, alongside the output from our algorithm. Each image was colorized in only 1 minute of time by a novice user. All of the user study results are publicly available on \url{https://richzhang.github.io/ideepcolor/}. Images are from the Imagenet dataset \cite{russakovsky2015imagenet}.}
\label{fig:userstudy_showcase}
\vspace{-2mm}
\end{figure*}

\subsection{Global Hints Network}
\label{sec:globalhintsnet}

% \subsubsection{System inputs}
An advantage of the end-to-end learning framework is that it may be easily adapted to different types of user inputs. We show an additional use case, where the user provides global statistics, described by a global histogram $\mathbf{X}_{hist}\in \Delta^{Q}$ and average image saturation $\mathbf{X}_{sat}\in [0,1]$. Whether or not the inputs are provided is indexed by indicator variables $\mathbf{B}_{hist},\mathbf{B}_{sat}\in \mathds{B}$, respectively. The user input to the system is then $\mathbf{U}_{g} = \{\mathbf{X}_{hist}, \mathbf{B}_{hist}, \mathbf{X}_{sat}, \mathbf{B}_{sat}\} \in \mathds{Re}^{1\times 1\times (Q+3)}$.

% \subsubsection{Training}
We compute global histograms by resizing the color $\mathbf{Y}$ to quarter resolution using bilinear interpolation, encoding each pixel in quantized $ab$ space, and averaging spatially. Saturation is computed by converting the ground truth image to HSV colorspace and averaging over the S channel spatially. We randomly reveal the ground truth colorization distribution, ground truth saturation, both, or neither, to the network during training.

\subsection{Network Architecture}
\label{sec:net-arch}

We show our network architecture in Figure \ref{fig:network}. The main colorization branch is used by both Local Hints and Global Hints networks. We then describe the layers which are only used for the Local Hints Network, namely processing the sparse user input $\mathbf{U}_l$ and the color distribution prediction branch, both shown in red. Finally, we describe the Global Hints Network-specific input branch, shown in green, as well as its integration in the main network.

\subsubsection{Main colorization network} The main branch of our network, $\mathcal{F}$, uses a U-Net architecture \cite{ronneberger2015u}, which has been shown to work well for a variety of conditional generation tasks \cite{isola2016image}. We also utilize design principles from \cite{simonyan2014very} and \cite{yu2015multi}. The network is formed by 10 convolutional blocks, \texttt{conv1-10}. In \texttt{conv1-4}, in every block, feature tensors are progressively halved spatially, while doubling in the feature dimension. Each block contains 2-3 \texttt{conv-relu} pairs. In the second half, \texttt{conv7-10}, spatial resolution is recovered, while feature dimensions are halved. In block \texttt{conv5-6}, instead of halving the spatial resolution, dilated convolutions with factor 2 is used. This has an equal effect on the receptive field of each unit with respect to the input pixels, but allows the network to keep additional information in the bottleneck. Symmetric shortcut connections are added to help the network recover spatial information \cite{ronneberger2015u}. For example, the \texttt{conv2} and \texttt{conv3} blocks are connected to the \texttt{conv8} and \texttt{conv9} blocks, respectively. This also enables easy accessibility to important low-level information for later layers; for example, the lightness value will limit the extent of the $ab$ gamut. Changes in spatial resolution are achieved using subsampling or upsampling operations, and each convolution uses a $3\times 3$ kernel. \texttt{BatchNorm} layers are added after each convolutional block, which has been shown to help training.

A subset of our network architecture, namely \texttt{conv1-8} without the shortcut connections, was used by Zhang et al.~\shortcite{zhang2016colorful}. For these layers, we fine-tune from these pre-trained weights. The added \texttt{conv9, conv10} layers and shortcut connections are trained from scratch. A last \texttt{conv} layer, which is a $1 \times 1$ kernel, maps between \texttt{conv10} and the output color. Because the $ab$ gamut is bounded, we add a final \texttt{tanh} layer on the output, as is common practice when generating images~\cite{goodfellow2014generative,zhu2016generative}.

\subsubsection{Local Hints Network} The layers specific to the Local Hints Network are shown in red in Figure~\ref{fig:network}. Sparse user points are integrated by concatenation with the input grayscale image. As a side task, we also predict a color \textit{distribution} at each pixel (conditioned on the grayscale and user points) to recommend to the user. The task of predicting a color distribution is undoubtedly related to the task of predicting a single colorization, so we reuse features from the main branch. We use a hypercolumn approach \cite{hariharan2015hypercolumns,larsson2016learning} by concatenating features from multiple layers of the main branch, and learning a two-layer classifier on top. Network $\mathcal{G}_l$ is composed of the main branch, up to \texttt{conv8}, along with this side branch. The side task should not affect the main task's representation, so we do not back-propagate the gradients from the side task into the main branch. To save computation, we predict the distribution at a quarter resolution, and apply bilinear upsampling to predict at full resolution.

\subsubsection{Global Hints Network} Because the global inputs have no spatial information, we choose to integrate the information into the middle of the main colorization network. As shown in the top green branch in Figure \ref{fig:network}, the inputs are processed through 4 \texttt{conv-relu} layers, with kernel size $1\times 1$ and 512 channels each. This feature map is repeated spatially to match the size of the \texttt{conv4} feature in the main branch, $\mathds{R}^{\sfrac{H}{8}\times \sfrac{W}{8}\times 512}$, and merged by summation, a similar strategy to the one used by Iizuka et al.~\shortcite{iizuka2016let}.

%% file: sections/4_experiments.tex
\begin{figure}[t!]
\includegraphics[width=1.0\hsize]{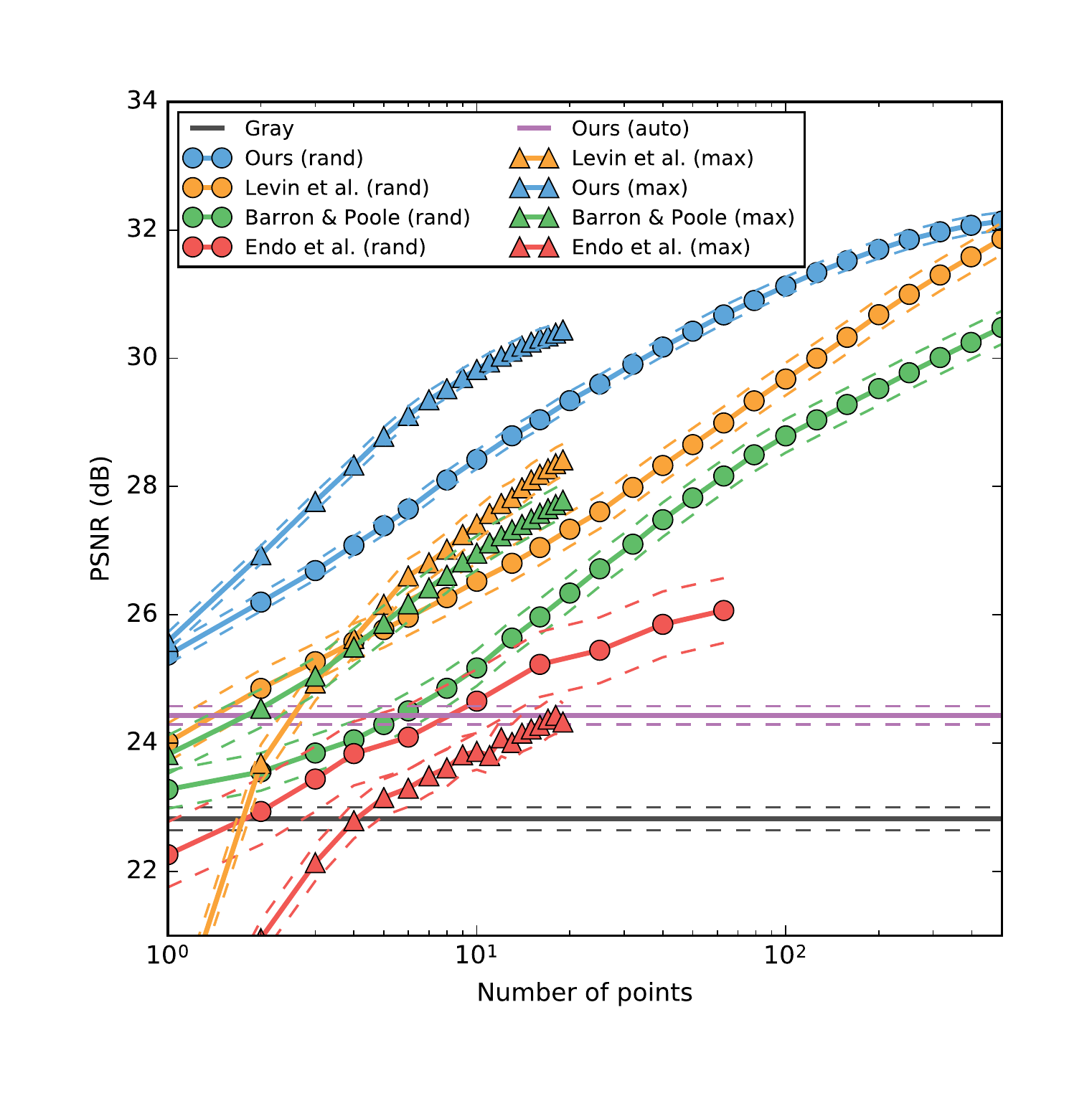}
\vspace{-6mm}
\caption{\textbf{Average PSNR vs Number of Revealed Points} We measure the average PSNR from our ImageNet test set across different algorithms. Points are revealed to each algorithm by \textbf{random} or \textbf{max}-error sampling. Max-error sampling selects the point with maximum $\ell_2$ error in $ab$ space between predicted and ground truth. Random sampling uses a uniformly drawn random point. The average color on a $7\times 7$ patch is revealed to the algorithm. The x-axis is on a logarithmic scale. Baselines~\cite{levin2004colorization,poole2016fast,endo2016deepprop} are computed with publicly available code from the authors. Because our algorithm is learned on a large-scale corpus of data, our system provides more accurate colorizations given little user supervision. With large amounts of input points (approximately 500 for random sampling), \cite{levin2004colorization} begins to achieve equal accuracy to ours. For reference, we show our network without user inputs, \textbf{Ours (auto)}, and predicting \textbf{Gray} for every pixel.}
\label{fig:psnr-points}
\vspace{-2mm}
\end{figure}

\section{Experiments}
\label{sec:expr}

We detail qualitative and quantitative experiments with our system. In Section \ref{sec:psnr}, we first automatically test the Local Hints Network.
%We iteratively reveal points from the ground truth image to the algorithm, and measure its ability to reconstruct the ground truth image.
We then describe our user study in Section \ref{sec:turk}. The results suggest that even with little training and just 1 minute to work with an image, novice users can quickly create realistic colorizations. In Section \ref{sec:unusual}, we show qualitative examples on unusual colorizations. In Section \ref{sec:expr_global} we evaluate our Global Hints Network. In Section \ref{sec:mult}, we investigate how the Local Hints Network reconciles two colors within a single segment. Finally, we show qualitative examples on legacy grayscale images in Section \ref{sec:legacy}.

\vspace{-2mm}
\input{tables/psnr_auto}

\subsection{How well does the system incorporate inputs?}
\label{sec:psnr}

We test the system automatically by randomly revealing patches to the algorithm, and measuring PSNR, as shown in Figure \ref{fig:psnr-points}. The pitfalls of using low-level or per-pixel metrics have been discussed in the automatic colorization regime \cite{zhang2016colorful}. A system which chooses a plausible but different mode than the ground truth color will be overly penalized, and may even achieve a lower score than an implausible but neutral colorization, such as predicting gray for every pixel (PSNR 22.8). In this context, however, since ground truth colors are revealed to the algorithm, the problem is much more constrained, and PSNR is a more appropriate metric.

With no revealed information, edit propagation methods will default to gray for the whole image. Our system will perform automatic colorization, and provide its best estimate (PSNR 24.4), as described in Table \ref{tab:psnr}. As points are revealed, PSNR incrementally increases across all methods. Our method achieves a higher PSNR than other methods, even up to 500 random points. As the number of points increases, edit propagation techniques such as \cite{levin2004colorization} approach our method, and will inevitably surpass it. In the limiting case, where every point is revealed, edit propagation techniques such as \cite{levin2004colorization,poole2016fast,endo2016deepprop} will correctly copy the inputs to the outputs (PSNR $\infty$). Our system is taught to do this, based on $1\%$ of the training examples, but will not do so perfectly (PSNR 37.70). As the number of points increases to the hundreds, knowledge of mid-to-high-level natural image statistics has diminishing importance, and the problem can be solved using low-level optimization.

We also run the same test, but with points sampled in a more intelligent manner. Given an oracle which provides the ground truth image, we compute the $\ell_2$ error between the current prediction and the ground truth, and average over a $25\times 25$ window. We then select the point with the maximum error to reveal a $7\times 7$ patch, excluding points which overlap with previously revealed patches. As expected, this sampling strategy typically achieves a higher PSNR, and the same trend holds -- our method achieves higher accuracy than the current state-of-the-art method. Inferring the full colorization of an image given sparsely revealed points has been previously exploited in the image compression literature \cite{cheng2007learning,he2009unified}. An interesting extension of our network would be to optimally choose which points to reveal.

We also note that our method has been designed with point inputs, whereas previous work has been designed with stroke and point-based inputs in mind. In an interactive setting, the collection cost of strokes versus points is difficult to define, and will heavily depend on factors such as proper optimization of the user interface. However, the results strongly suggest that our method is able to accurately propagate sparse, point-based inputs.

\subsection{Does our system aid the user in generating realistic colorizations?}
\label{sec:turk}

\input{tables/turk_table}

We run a user study, with the goal of evaluating if novice users, given little training, can quickly produce realistic colorizations using our system. We provide minimal training for 28 test subjects, briefly walking them through our interface for 2 minutes. The subjects are given the goal of producing ``realistic colorizations" (without benefit of seeing the ground truth), and are provided 1 minute for each image. Images are randomly drawn from our ImageNet test set. Each subject is given 20 images -- 10 images with our algorithm and full interface, including suggested colors, and 10 images with our algorithm but no color suggestions, for a total of 560 images (280 per test setting). We evaluate the resulting colorizations, along with automatic colorization, by running a real vs. fake test on Amazon Mechanical Turk (AMT), using the procedure proposed by Zhang et al.~\shortcite{zhang2016colorful}. AMT evaluators are shown two images in succession for 1 second each -- one ground truth and one synthesized -- and asked to identify the synthesized. We measure the ``fooling rate" of each algorithm; one which produces ground truth colorizations every time would achieve $50\%$ by this metric. The results are shown in Table \ref{tab:turk}. Note that the results may differ on an absolute scale from previous iterations of this test procedure \cite{zhang2016colorful,isola2016image}, due to shifts or biases in the AMT population when the algorithm has been tested. Our network produces a fooling rate of $18.6\%$ when run completely automatically (no user inputs). We test our interface without recommended colors, but with HSV sliders and 48 common colors. With this baseline interface, the fooling rate increases dramatically to $27.0\%$, indicating that users quickly acclimated to our network and made dramatic improvements with just 1 minute. When provided the data-driven color palette, the fooling rate further increases to $30.0\%$. This suggests that the color prediction feature can aid users in quickly selecting a desired color.

We show example results from our study in Figures \ref{fig:userstudy_res} and \ref{fig:userstudy_showcase}. We compare the annotations to the seminal method proposed by Levin et al.~\shortcite{levin2004colorization}, along with the automatic output from our network. Qualitatively, the added user points typically add (1) saturation when the automatic result is lacking and (2) accurate higher frequency detail, that automatic methods have difficulty producing. Comparing our method to Levin et al.~\shortcite{levin2004colorization}, our method is more effective at finding segment boundaries given sparse user inputs. We do note that the user points are collected by running our system, which provides an advantage. However, collecting these points, with the right colors, is enabled by the interactive nature of our algorithm and our color recommendation system.

\begin{figure}[t!]
    \begin{subfigure}[t]{0.15\textwidth}
        \centering
        \includegraphics[width=\textwidth]{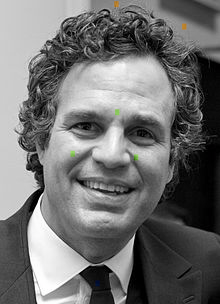}
        \caption{}
    \end{subfigure}
    \begin{subfigure}[t]{0.15\textwidth}
        \centering
        \includegraphics[width=\textwidth]{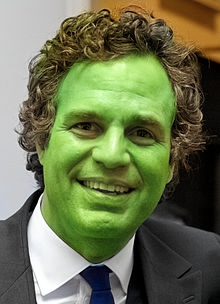}    
        \caption{}
    \end{subfigure}
    \begin{subfigure}[t]{0.15\textwidth}
        \centering
        \includegraphics[width=\textwidth]{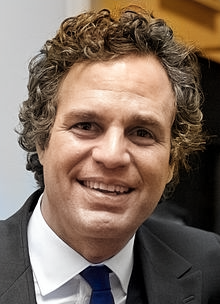}
        \caption{}
    \end{subfigure}
    \vspace{-3mm}
\caption{\textbf{Unusual colorization} \textbf{(a)} User inputs with unusual colors \textbf{(b)} Output colorization using user points with unusual colors \textbf{(c)} Output colorization with user points using conventional colors. Photograph by Corporal Michael Guinto, 2014 (Public Domain).
}
\label{fig:unusual}
\vspace{-4mm}
\end{figure}

\begin{figure}[t!]
    \centering
    \begin{subfigure}[t]{0.075\textwidth}
        \centering
        \includegraphics[width=\textwidth]{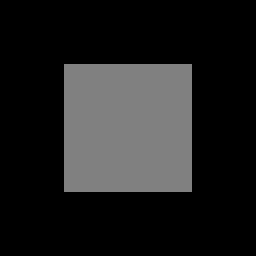}
        \caption{}
    \end{subfigure}
    \begin{subfigure}[t]{0.075\textwidth}
        \centering
        \includegraphics[width=\textwidth]{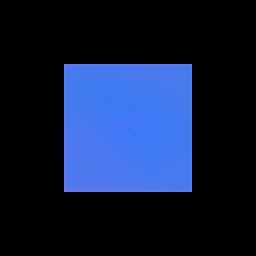}
        \caption{}
    \end{subfigure}
    \begin{subfigure}[t]{0.075\textwidth}
        \centering
        \includegraphics[width=\textwidth]{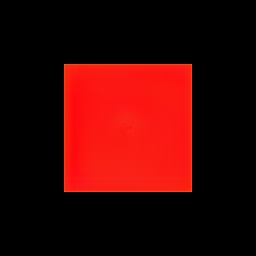}    
        \caption{}
    \end{subfigure}
    \begin{subfigure}[t]{0.075\textwidth}
        \centering
        \includegraphics[width=\textwidth]{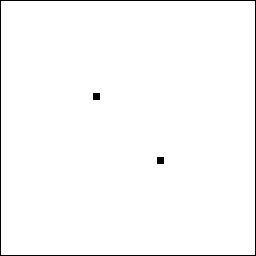}
        \caption{}
    \end{subfigure}
    \begin{subfigure}[t]{0.075\textwidth}
        \centering
        \includegraphics[width=\textwidth]{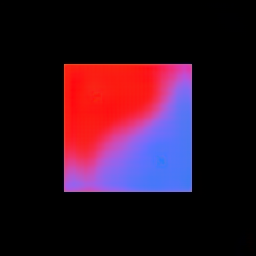}  
        \caption{}
    \end{subfigure}
    \begin{subfigure}[t]{0.075\textwidth}
        \centering
        \includegraphics[width=\textwidth]{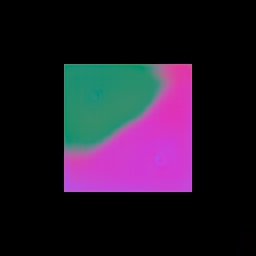}  
        \caption{}
    \end{subfigure}
    \vspace{-3mm}
    \caption{\textbf{Multiple user colors within a segment.} \textbf{(a)} Input grayscale image. \textbf{(b,c)} Output colorization conditioned on a single centered user point colored (b-blue, c-red). \textbf{(d)} Locations used for user points for (e) and (f). \textbf{(e,f)} Outputs given different user input colors (e-blue\&red, f-green\&pink).}
    \label{fig:mult}
    \vspace{-6mm}
\end{figure}

\begin{figure*}[t!]
\includegraphics[width=1.0\hsize]{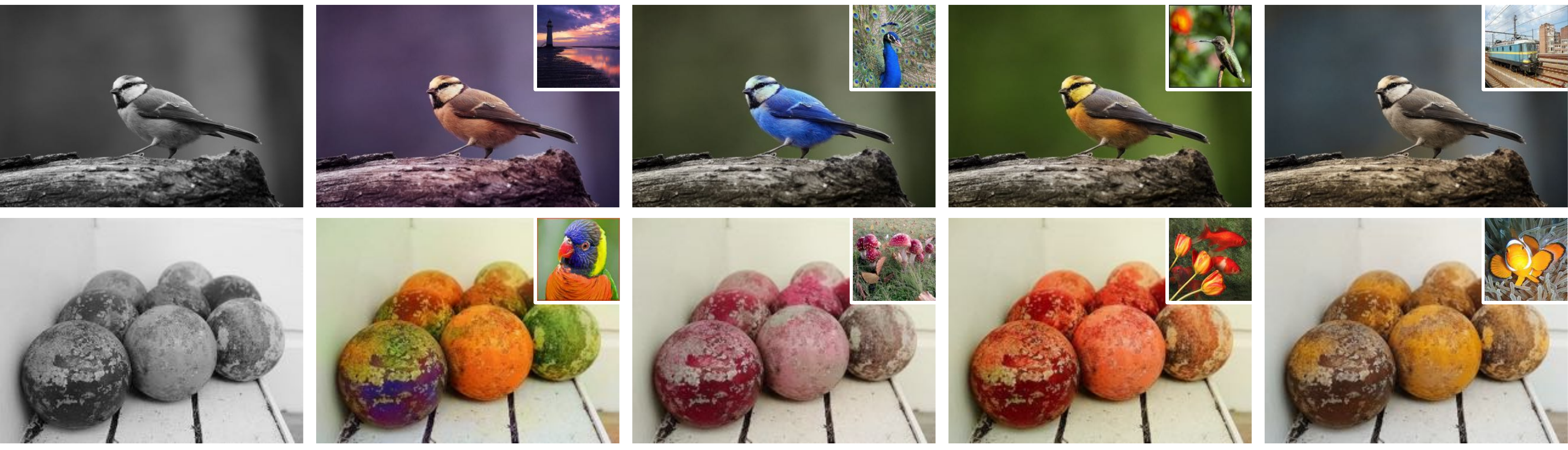}
\vspace{-6mm}
\caption{\textbf{Global histogram transfer} Using our Global Hints Network, we colorize the grayscale version of the image on the left using global histograms from the top-right inset images. Images are from the Imagenet dataset \cite{russakovsky2015imagenet}.}
\label{fig:bird_figure}
% \vspace{-2mm}
\end{figure*}

\begin{figure*}[t!]
\includegraphics[width=1.\hsize]{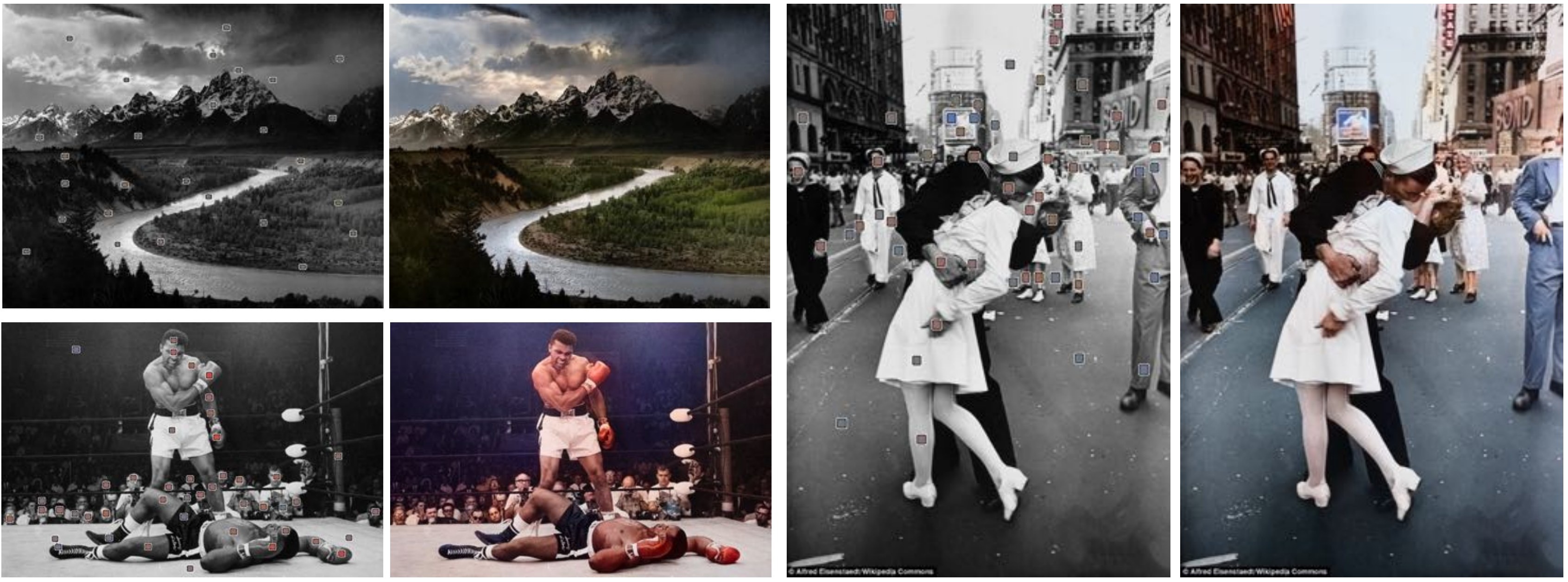}
\vspace{-6mm}
\caption{\textbf{Legacy black and white photographs} Our method applied to legacy black and white photographs. Top left: \emph{The Tetons and Snake River}, Ansel Adams, 1942; Bottom left: Photo by John Rooney of Muhammad Ali versus Sonny Liston, 1965 (c.f. color photo by Neil Leifer at almost exactly the same moment); Right: \emph{V-J Day in Times Square}, Alfred Eisenstaedt, 1945.
}
\label{fig:legacy_res}
% \vspace{-3mm}
\end{figure*}

\subsection{Does the network generalize to unusual colorizations?}
\label{sec:unusual}

During training, we use natural images, and reveal the \textit{ground truth colors} to simulate user input. However, there are use cases where the user may intentionally desire an unusual colorization. Will the network be able to follow the inputs in these cases? In Figure \ref{fig:unusual}, we show an unusual colorization guided by the user, giving the actor a green face with three user points on the face. These results suggest that in the absence of nearby user inputs, the network will attempt to find an appropriate colorization for the object, based on the training corpus. However, once an input is provided by the user, the system fills in the segment with the desired color.

\vspace{-2mm}
\subsection{Is the system able to incorporate global statistics?}
\label{sec:expr_global}

We train a variant of our system, taking global statistics as inputs, instead of local points. As described in Table \ref{tab:psnr}, when given the ground truth statistics, such as the global histogram of colors or average saturation, the network achieves a higher PSNR scores, 27.9 and 25.6, respectively, than when performing automatic colorization (24.4), indicating that the network has learned how to fuse global statistics. We also test on the SUN-6 dataset, shown in Table \ref{tab:sun6}, proposed by Deshpande et al.~\shortcite{deshpande2015learning}. We show higher performance than Despande et al.~\shortcite{deshpande2015learning} and almost equal performance with Larsson et al.~\shortcite{larsson2016learning}, which fuses the predictions from an automatic colorization network with a ground truth histogram using an energy minimization procedure.

The network has only been trained on images with its own ground truth histogram. In Figure \ref{fig:bird_figure}, we qualitatively the network's generalization ability by computing the global histogram on separate reference images, and apply them to a photograph. The bird is an interesting test case, as it can be plausibly colorized in many different ways. We observe that that the color distributions of the reference input image is successfully transferred to the target grayscale image. Furthermore, the colorizations are realistic and diverse.

\input{tables/sun6_table}

\subsection{How does the system respond to multiple colors within an equiluminant segment?}
\label{sec:mult}

In natural images, chrominance changes almost never appear without a lightness change. In Figure \ref{fig:mult}(a), we show a toy example of an image of a gray square on top of a black square. If given a $7\times 7$ point in the center of the image, the system will successfully propagate the color to the center region, as shown in Figures \ref{fig:mult}(a)(b). However, how does the system respond if given two different colors within the same segment, as shown in Figure \ref{fig:mult}(d)? Given blue and red points, the system draws a seam between the two colors, as shown in Figure \ref{fig:mult}(e), where two points are placed symmetrically around the center of the image. Because our system is learned from data, it is difficult to characterize how the system will exactly behave in such a scenario. Qualitatively, we observe that the seam is not straight, and the shape as well as the sharpness of the transition is dependent on the colors. For example, in Figure \ref{fig:mult}(f), green and pink points produce a harder seam. We found similar behavior under similar scenarios in natural images as well.

\subsection{Is the system able to colorize legacy photographs?}
\label{sec:legacy}

Our system was trained on ``synthetic" grayscale images by removing the chrominance channels from color images. We qualitatively test our system on \textit{legacy} grayscale images, and show some selected results in Figure \ref{fig:legacy_res}.

%% file: tables/psnr_auto.tex
\begin{table}[t!]
\begin{center}
\begin{tabular}{ l c c }
\specialrule{.1em}{.1em}{.1em}
\textbf{Method} & \textbf{Added Inputs} & \textbf{PSNR (dB)} \\ \hline
\specialrule{.1em}{.1em}{.1em}
Predict gray & -- & 22.82$\pm$0.18 \\ \hline
Zhang et al.~\shortcite{zhang2016colorful} & automatic & 22.04$\pm$0.11 \\ 
Zhang et al.~\shortcite{zhang2016colorful} (no-rebal) & automatic & 24.51$\pm$0.15 \\ 
% \cite{larsson2016learning} & automatic & 24.02$\pm$0.15 \\ % running at 256x256
% \cite{larsson2016learning} & automatic & 24.84$\pm$0.14 \\ % running at full resolution
Larsson et al.~\shortcite{larsson2016learning} & automatic & 24.93$\pm$0.14 \\ % running at full resolution, resize to 256x256
Iizuka et al.~\shortcite{iizuka2016let} & automatic & 23.69$\pm$0.13 \\ 
Ours (Local) & automatic & 24.43$\pm$0.14 \\ \hline
Ours (Global) & + global hist & 27.85$\pm$0.13 \\ % 45k
Ours (Global) & + global sat & 25.78$\pm$0.15 \\ % 45k
Ours (Local) & + gt colors & 37.70$\pm$0.14 \\ \hline
Edit propagation & + gt colors & $\infty$ \\
\specialrule{.1em}{.1em}{.1em}
\end{tabular}
\end{center}
\caption{\textbf{PSNR with added information.} Run on 1000 held-out test images, in the ILSVRC2012 \cite{russakovsky2015imagenet} validation dataset. \textbf{Ours (Local)-automatic} is run completely automatically, with no user inputs. Methods \cite{zhang2016colorful,larsson2016learning,iizuka2016let} are recently automatic colorization methods. Even though our network is trained primarily for interactive colorization, it performs competitively for automatic colorization as well by this metric. \textbf{Ours (Global) +global hist} provides global distribution of colors in the $ab$ gamut; \textbf{Ours (Global) +global sat} provides global saturation to the system. Our Global Hints Network learns to incorporate global statistics for more accurate colorizations. %\textbf{Ours (Local) + gt colors} uses ground truth colors. Given ground truth colors, our network
}
\vspace{-4mm}
\label{tab:psnr}
\end{table}

%% file: tables/turk_table.tex
\begin{table}[t!]
\begin{center}
\begin{tabular}{ l c }
\specialrule{.1em}{.1em}{.1em}
\textbf{Method} & \textbf{AMT Fooling Rate} \\ \hline
\specialrule{.1em}{.1em}{.1em}
Ours-automatic & 18.58\% $\pm$ 1.09 \\
Ours-no recommendation & 26.98\% $\pm$ 1.76 \\
Ours & \textbf{30.04\% $\pm$ 1.80} \\
\specialrule{.1em}{.1em}{.1em}
\end{tabular}
\end{center}
\caption{\textbf{Amazon Mechanical Turk real vs fake fooling rate} We test how often colorizations generated by novice users fool real humans. \textbf{Ours} is our full method, with color recommendations. \textbf{Ours-no recommendation} is our method, without the color recommendation system. \textbf{Ours-automatic} is our method with no user inputs. Note that the 95\% confidence interval shown is not accounting for possible inter-subject variation (all subjects are assumed to be identical).}
\label{tab:turk}
\vspace{-10mm}
\end{table}

%% file: tables/sun6_table.tex
\begin{table}[t!]
\begin{center}
\begin{tabular}{ l c c }
\specialrule{.1em}{.1em}{.1em}
\textbf{Method} & \textbf{Added Inputs} & \textbf{PSNR (dB)} \\ \hline
\specialrule{.1em}{.1em}{.1em}

Deshpande et al.~\shortcite{deshpande2015learning} & automatic & 23.18 $\pm$ 0.20 \\
Larsson et al.~\shortcite{larsson2016learning} & automatic & 25.60 $\pm$ 0.23 \\
Ours & automatic & 25.65 $\pm$ 0.23 \\ \hline % saturation with 45k

Deshpande et al.~\shortcite{deshpande2015learning} & + global hist & 23.85 $\pm$ 0.23 \\
Larsson et al.~\shortcite{larsson2016learning} & + global hist & 28.62 $\pm$ 0.23 \\
Ours & + global hist & 28.57 $\pm$ 0.21 \\ % saturation with 45k

\specialrule{.1em}{.1em}{.1em}
\end{tabular}
\end{center}
\caption{\textbf{Global Histogram} We test our Global Hints Network at incorporating the global truth histogram on 240 images from SUN used by \cite{deshpande2015learning}.}
\label{tab:sun6}
\vspace{-10mm}
\end{table}

%% file: sections/5_conclusions.tex
\section{Limitations and Discussion}

A benefit of our system is that the network predicts user-intended actions based on learned semantic similarities. However, the network can also be over-optimistic and produce undesired non-local effects. For example, points added on a foreground object may cause an undesired change in the background, as shown on the last row in Figure \ref{fig:userstudy_res}. Qualitatively, we found that adding some control points can remedy this. In addition, the network can also fail to completely propagate a user point, as shown in the fifth row in Figure \ref{fig:userstudy_res}. In these instances, the user can fill in the region with additional input.

For scenes with difficult segmentation boundaries, the user sometimes needs to define boundaries explicitly by densely marking either side. Our system can continuously incorporate this information, even with hundreds of input points, as shown on Figure \ref{fig:psnr-points}. Points can be added to fix color bleeding artifacts when the system has poor underlying segmentation. However, our interface is mainly designed for the ``few seconds to couple minutes" interaction regime. For users wanting high-precision control and willing to spend hours per photograph, working in Photoshop is likely a better solution.

Our system is currently trained on points; we find that in this regime random sampling covers the low-dimensional workspace surprisingly well. However, a future step is to better simulate the user, and to effectively incorporate stroke-based inputs that traditional methods utilize. Integration between the local user points and global statistics inputs would be an interesting next step. Our interface code and models are publicly available at \url{https://richzhang.github.io/ideepcolor}, along with all images generated from the user study and random global histogram transfer results.

%% file: sections/6_acknowledgements.tex
\section*{Acknowledgements}

% \noindent \textbf{Acknowledgements}
We thank members of the Berkeley Artificial Intelligence Research Lab for helpful discussions. We also thank the participants in our user study, along with Aditya Deshpande and Gustav Larsson for providing images for comparison. This work has been supported, in part, by NSF SMA-1514512, a Google Grant, BAIR, and a hardware donation by NVIDIA.
% \vspace{-2mm}

%% file: sections/7_changelog.tex
\section*{Change Log}

% \noindent \textbf{Change Log} 
\noindent \textbf{v1} Initial release. SIGGRAPH camera ready version. DOI: \url{http://dx.doi.org/10.1145/3072959.3073703}
% \vspace{-2mm}